# Student Engagement Detection Using Emotion Analysis, Eye Tracking and Head Movement with Machine Learning

Prabin sharma [1], Shubham Joshi [2], Subash Gautam [2] , Sneha Maharjan [3] , Salik Ram Khanal [4] , Manuel Cabral Reis [5,6] João Barroso [5,7] and Vítor Manuel de Jesus Filipe [5,7]

[1] University of Massachusetts Boston, USA
[2] Kathmandu University, Dhulikhel, Nepal
[3] Wentworth institute of technology, Boston, USA
[4] Center for Precision and Automated Agricultural Systems, Washington State University, Prosser, USA
[5] University of Trás-os-Montes e Alto Douro Vila Real, Portugal
[6] Institute of Electronics and Informatics Engineering of Aveiro, Portugal
[7] INESC TEC

**Abstract.** With the increase of distance learning, in general, and e-learning, in particular, having a system capable of determining the engagement of students is of primordial importance, and one of the biggest challenges, both for teachers, researchers and policy makers. Here, we present a system to detect the engagement level of the students. It uses only information provided by the typical built-in web-camera present in a laptop computer, and was designed to work in real time. We combine information about the movements of the eyes and head, and facial emotions to produce a concentration index with three classes of engagement: “very engaged”, “nominally engaged” and “not engaged at all”. The system was tested in a typical e-learning scenario, and the results show that it correctly identifies each period of time where students were “very engaged”, “nominally engaged” and “not engaged at all”. Additionally, the results also show that the students with best scores also have higher concentration indexes.

## 1 Introduction

Students of the 21st century are moving to a Digital Education, focusing on teachers and students’ relations to achieve the goal of meaningful, high quality and dynamic education. The advent of digitization in education has brought drastic changes in education system. However, there are still some challenges that teachers/instructors are facing. As mentioned by [1], one of the challenges that the teachers/instructors are facing is to examine how well the students/learner are receiving the content delivered from the lecture. Student engagement, which occurs when the student involves meaningfully through the learning environment, is a topic of paramount importance and should be taken carefully in order to improve the educational system. As defined by [2], student engagement is the psychological investment of student in learning and understanding the knowledge, skills or crafts that academic work is trying to encourage. Engagement is directly proportional to student's achievement [3].

The concept of Virtual Classroom was implemented for the first time in mid 1990s [4]. At the same time, the World Wide Web has become a popular way to deliver the content to students. As a natural consequence, virtual classroom systems have been adapted in many schools. However, one of the most important problems of virtual classroom systems is the dropout rate of students.

The problem of disengagement of students is being raising attention every day. Low achievement of the student is a relative minor problem when compared to disengagement of students (and is a direct consequence of disengagement). Reyes et al. [5] have collected data from 63 fifth- and sixth grade classrooms (N = 1,399 students) and they have found that good emotional climate and grade are mediated by engagement. As stated by Stanley and Hanse [6], students not paying attention in class is one of the main indicators of disengagement. Their degree of attention and curiosity reflects their engagement in class. Psychological effect and socio-cultural orientation that students bring to school can be the external factors that lead to the declination in degree of attention. Bradbury [7] concluded that the way of teaching is also a reason behind engagement of students; they reported that between 25% to 60% of the students were bored for a long period of time and disengaged in the classroom environment [8]. Ekman, Friesen, and Ellsworth, [9], state that the fast way to understand emotion is with the help of facial expression that people express. Emotions of students during their learning period (in the classroom or any other learning environment) can be used as a useful information to evaluate their concentration towards the "delivered" content. In particular, eye and head movements can be used to determine the engagement of students when using a computer (e.g., in virtually classes). These movements can be used to estimate how much students are concerned with the delivered contents. For example, eye tracking is being used to evaluate human behaviors and predict the degree of attention [10]. As stated by Poole and Ball [11], eye tracking is the system in which a person's eyes movements are measured, letting the researcher know where the person is looking, at any given time, and the sequence in which the person's eyes are moving from one position to another. Both eye tracking hardware and software algorithms can be used to extract the information from the movement of eyes (or only one eye) [12], [13]. Other authors suggest the use of other data, such as pupil dilation, because it occurs when the students see emotional arousal pictures [14], or eye closure duration [1], to identify the engagement of students.

In a typical e-learning environment, students have a laptop computer with a built-in camera. Here, we propose the use of the laptop built-in web-cam to grab real time information about the eye's movements (eye tracking) and facial emotions of the students. This information will be used to determine a concentration level, hence helping the instructor to see how engaged (or not) the students are. We believe that this information will help the teacher in making the learning environment affordable. To produce the concentration index presented here, in real-time, Python and Keras for the facial emotion analysis, the Haar-cascade algorithm for the eye tracking, and a Convolution Neural Network (CNN) are used.

## 2 Related Works

The problem of finding indexes to determine the concentration and engagement of students is being gaining attention in the recent years. These indexes can be of particular usefulness when the students are using autonomous e-learning systems, where no teacher/instructor is present and so the feedback about the reactions, emotions, etc., of the students are not easy to grab. The number of researches addressing these problems is being growing every day.

The research conducted by Divjak and Bischof [15] analysed and evaluated three variables (eye tracking, head movement and eye close duration) to produce an alert when they find the user having "computer vision syndrome". They used Open CV to localize the head and eye and set the threshold value for both movements, if the movement crosses the minimum threshold value, the alert will be generated to notify the user.

A research work done by Turabzadeh et al. [16] was based on facial emotion recognition in real-time, using the Local Binary Point (LBP) algorithm, in which LBP features were extracted from the video captured, which was then used as input for a K-Nearest Neighbour (K-NN) regression with dimensional labels. The system's accuracy, using MATLAB Simulink, reached 51.28% and in the Xilinx simulation was 47.44%.

Bidwell and Fuchs [17] measured students' engagement with an automated gaze system. They designed a student engagement classifier by using recorded video in classes. They used a face tracking system to extract students' gaze. The resulting automated gaze pattern was correlated with the pattern produced by a panel of experts' observations, for the training of a Hidden Markov Model (HMM). However, HMM resulted in a poor classification; they proposed to produce 8 discrete behaviours categories, but only were able to classify weather a student is "engaged" or "not engaged".

Krithika [18] uses eye and head movements for checking the concentration of students and generate a low concentration alert. The video was divided into frames and then taken into analysis. The implementation was done in MATLAB, using different functions for face detection and the ViolasJones features detection. The system is efficient enough to detect the negative emotions of the student in e- learning environments.

Kamath, Biswas, and Balasubramania [19] use the VilolaJones face detection algorithm for the analysis of the input images, and then Histogram of Oriented Gradients (HOG), for the facial representation for the patch to get the final vector of features. Those features were used to train the instance-weighted Multiple Kernel Learning-Support-Vector Machine (MKL-SVM) to build a model and then the performance of the system was measured. They reached an average accuracy of 43.98%, and a maximum accuracy of 50.77%.

Sharma et al. [20] proposed a real time system, based on expressed facial emotions during a lesson, to check the students' concentration in an e-learning context, automatically adapting the contents according to the student's concentration level, by analyzing the student's emotions. The emotions are processed to find the final concentration index. The results have proved that the emotions expressed were correlated with the concentration of the students, and devised three distinct levels of concentration (high, medium, and low).

Deep learning was utilized by Omid [21] to train a face expression identification model using pre-existing image data, and they used the models weight to initialize a new engagement recognition model based to recognize engagement. 4627 engaged and disengaged image datasets were used to train the new model. When they used deep learning architectures to engagement recognition for the first time, they discovered that the engagement model outperformed them.

Sheng [22] suggested a novel paradigm for measuring learning engagement that uses facial expression recognition to quickly track learners' changing emotional states. Additionally, a brand-new facial expression detection technique based on domain adaptation is put forth that is appropriate for the MOOC scenario. The results of the experiments demonstrate effectiveness of their suggested framework to measure students' participation in their learning. They compared their framework with the state-of-art methods which showed superiority of their proposed framework.

Islam [23] designed a methodology for measuring students' levels of participation in traditional classroom settings as well as online learning environments. The suggested framework records the user's video and follows the user's face as it moves through the frame of the video. From the user's face, several features are extracted, including learnt features, learned facial fiducial points, head attitude, eye stare, etc. The Facial Action Coding System (FACS), which breaks down facial expressions into the basic movements of certain muscles or groups of muscles, is then utilized to detect these aspects (i.e., action units). The student's willingness to participate in the learning process (also known as behavioral engagement) and his or her emotional attitude toward learning are subsequently measured using the decoded action units (AUs) (i.e., emotional engagement). This system will enable the professor to get immediate feedback from body kinesics such facial expressions and gaze.

## 3 System Architecture

Fig. 1. presents the main blocks of the system proposed for student's engagement evaluation. The system can be used during the real time lecture session, whether in a virtual

class or remote learning scenario, or in any other learning environment using a laptop computer with a built-in web camera.

In the learning environment there are two participants: the instructor and the learner. When the student is interacting with the learning material, the image data of learner (captured by the web-camera) are automatically analyzed by the system to evaluate the student's concentration level. If the resulting concentration index falls below a pre-defined threshold value, an alert will be issued (e.g., to the instructor/teacher or to the learner itself).

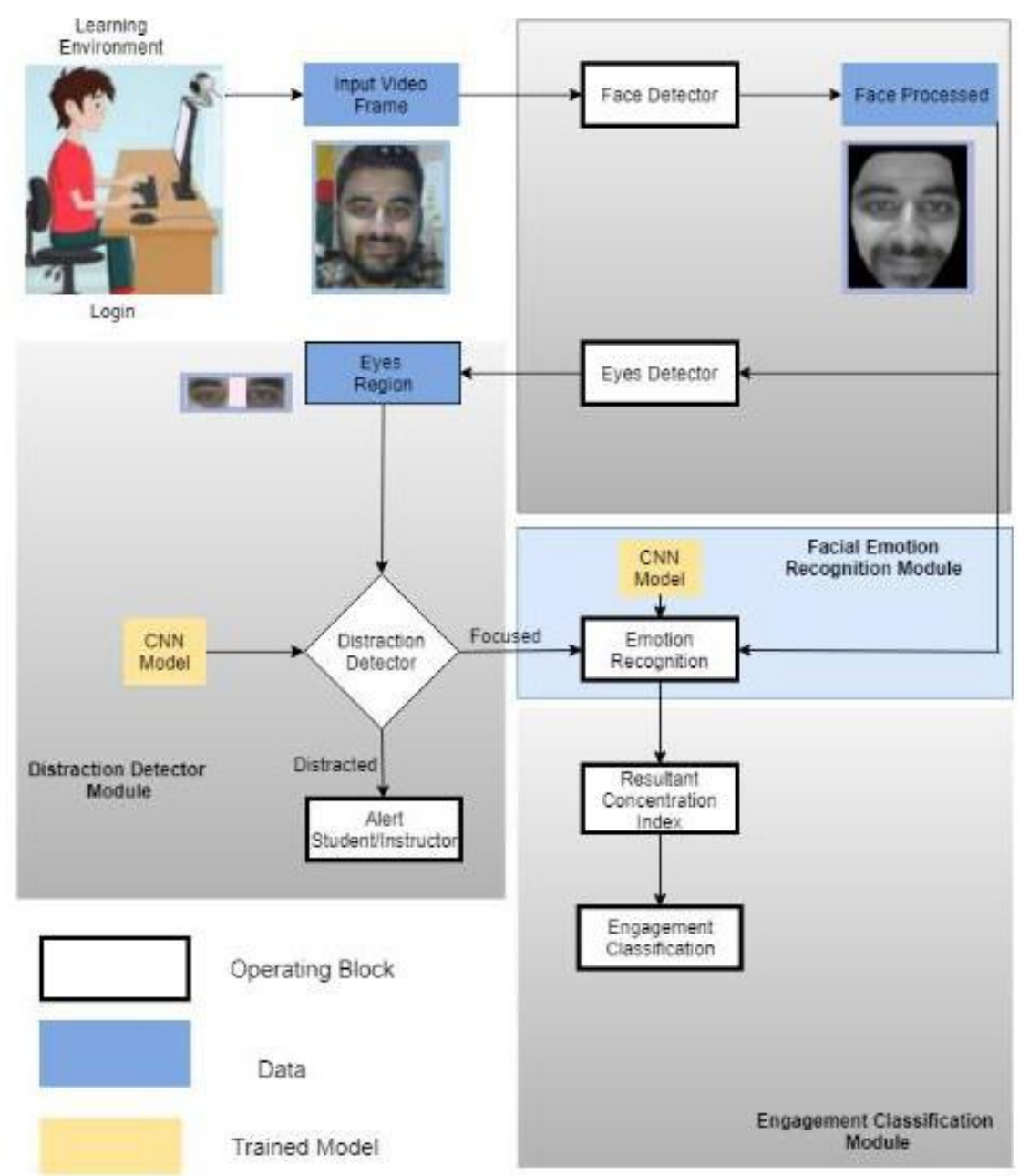

**Fig. 1.** Proposed system for student's engagement evaluation.

The detector engagement system comprises three modules, as shown in Fig .1. :

**a) Distraction Detector**

In each video frame the student's face is detected using the Viola & Jones algorithm. Next, within the detected face, the eyes region is located. The eyes region feeds a Convolutional Neural Network (CNN), used as a binary classifier, to predict the student's attention state in the two categories "Distracted" or "Focused".

**b) Facial Emotion Recognition**

Only when the student is "Focused", further facial emotions analysis will take place. For this purpose, another CNN model recognizes the dominant emotion expressed by the student's face at each moment. The classification is based on the emotion shown in the facial expression which can be one of the seven categories: Angry, Disgust, Fear, Happy, Sad, Surprise or Neutral.

**c) Engagement Classification**

The concentration index is calculated using the confidence score of dominant emotion and emotion weights. The resultant concentration index, a score between 0% and 100%, is used to classify the student's level of engagement in one of three categories: Very Engaged, Nominally Engaged and Not Engaged.

The system operates according to the following main steps:
Step 1: The student logs into the learning environment and the camera starts image acquisition.
Step 2: The face is detected and processed.
Step 3: The eyes region is detected and cropped.
Step 4: The student's attention state is classified in "Distracted" or "Focused";
Step 5: If the student is focused, dominant facial emotion is recognized.
Step 6: The resultant concentration index is calculated based on the confidence value and respective concentration index of dominant emotion.
Step 7: Finally, the student's engagement level is determined.

Two relevant algorithms are used in the detector engagement system: the Haar Cascade Algorithm; and a Convolution Neural Network.

**a) Haar Cascade Algorithm**

The Viola & Jones object detection algorithm uses the so-called Haar Cascade algorithm to extract features from images in a rapid and efficient way, and it had become one of the most popular methods currently being used to this end [21]. It needs a lot of positive and negative images to be trained for the cascade function to work properly. After getting trained, it can detect other images according to the previous training.

In this work we use this algorithm to detect the student's frontal face in the image and locate the eyes region within the face.

**b) Convolution Neural Network (CNN)**

CNN are distinct from traditional Artificial Neural Networks because they have the ability to encode relevant image features directly from the raw input images, making them more efficient to implement and reducing the number of parameters in the network [22].

In this work a CNN was trained on eye images to detect if the student is facing the webcam ("Focused") or not ("Distracted"), performing a binary classification in these two categories.

The CNN architecture comprises the following layers:

- Input layer 64x64 to hold the raw pixel values of the image;
- Convolutional layers with a set of 3 × 3 filters, to compute the output of neurons connected to local regions in the image.
- Pooling layer of 2×2 to reduce the spatial size of data representation.
- Fully connected layer to compute the two classes scores.

A second CNN, based on Arriaga, Valdenegro-Toro, and Ploger [23] work, was trained with grayscale images to ¨ classify facial emotions belonging to one of the following classes "angry", "disgust", "fear", "happy", "sad", "surprise", "neutral". This classification CNN model architecture, named mini-Xception which was inspired by Xception, developed by Franc¸ois Chollet [24], is a fully-convolutional neural network that contains Conv2D, residual depth-wise and separable Conv2D layer, each one followed by a batch normalization operation and a ReLU activation function. Finally, the output layer predicts the probabilities of seven emotions. The emotion with the highest probability score is considered the dominant emotion. When tested in the FER2013 dataset this architecture obtained an accuracy of 66% in emotion classification.

To train this CNN, we used the data-set from Kaggle challenge, which consists of 48 × 48 pixels gray scale images of faces and 35,887 examples. Fig. 2 presents the achieved accuracy, depending on the limited data-set and computational power. As can be seen, the accuracy increment is directly proportional to the number of epochs (one epoch means one pass through the full training set), but the accuracy remains constant after the 81st epoch, which means that the accuracy does not increase after that period or change insignificantly.

### c) Concentration Index

Initially, for the eye/head movement, we used the Haar cascade algorithm, which gives the result as a binary classifier ("Distracted" or "Focused"). Only when the student is "Focused", further facial emotions analysis will take place. CNN classifies the facial emotions data, generating the Dominant Emotion Probability (DEP) score. We have used the seven emotions that basically a person expresses: Neutral, Happy, Surprise, Sad, Disgust, Anger, and Scare. Table 1 represents Dominant Emotions and their corresponding Weights. The resulting Concentration Index (CI) is determined by multiplying the Dominant Emotion Probability (DEP) value by the corresponding Emotion Weight (EW), according to Table 1, expressed by equation 1:

$$\text{CI} = \text{DEP} \times \text{EW} \tag{1}$$

**Table 1.** Dominant Emotions and their corresponding Weights

| Dominant Emotion | Emotion Weight |
|---|---|
| Neutral | 0.9 |
| Happy | 0.6 |
| Surprised | 0.6 |
| Sad | 0.3 |
| Disgust | 0.2 |
| Anger | 0.25 |
| Scared | 0.3 |

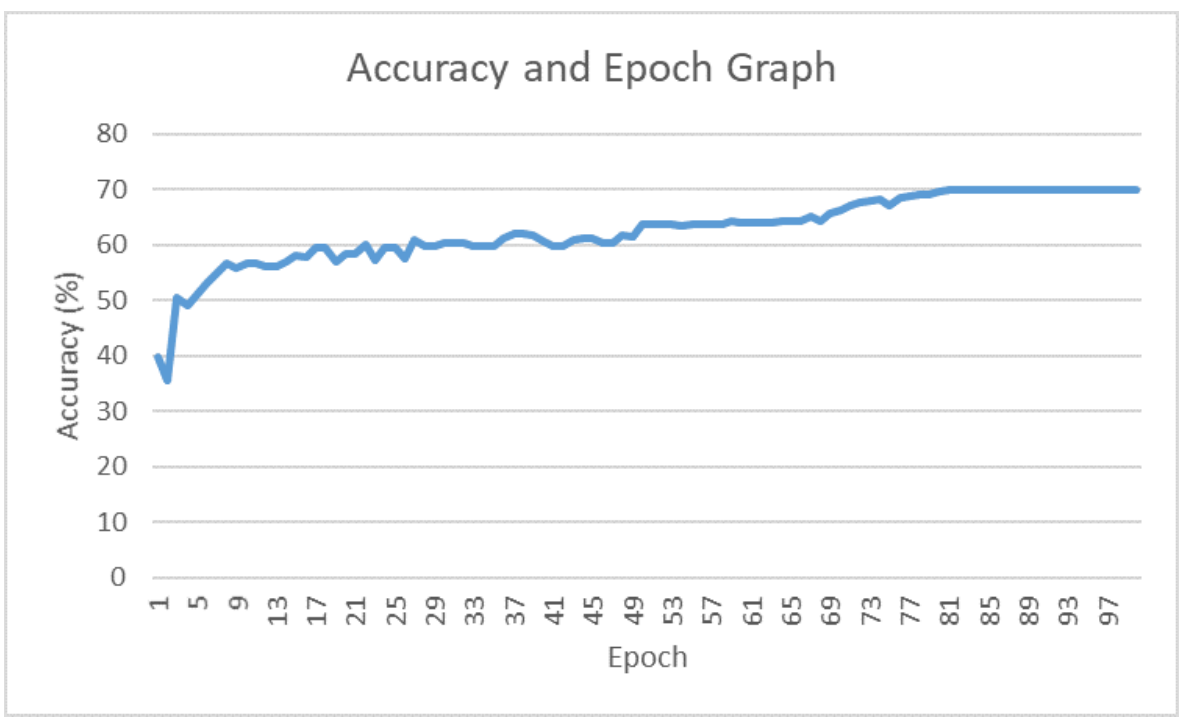

**Fig. 2.** Evolution of the accuracy as a function of epoch.

Emotion weight is defined as the value that describes how much a specific emotion state reflects the concentration of a student at that point of time. The value ranges from 0 to 1. To get the weights corresponding to each emotion, an informative video was shown to 30 students followed by a quiz with 10 questions. Data of facial emotion was recorded for all the students. Students were grouped on the basis of their major emotion expressed; for example, if a student expressed the majority of time (more than 50% of the time video duration) a neutral expression, he/she was included in the neutral emotional group. The students were distributed among the seven emotional groups according to the facial expressions obtained during the video. The mean score achieved in the quiz for each group was calculated and is presented in Table 1. Hence, the score of each group at some extent reflects the relation between the concentration of a student and its emotional state. Fig. 2. Represents the evolution of the accuracy as a function of epoch.

#### d) Categories of engagement

We will have two outputs, one resulting from the analysis of the movements of the eyes and head, and one resulting from the facial emotions analysis. By analyzing both components, presented in the results section, we decided to divide the engagement level into three different categories: very engaged; nominally engaged; and not engaged at all, as described below.

- Very engaged: a student engagement is under this category when its concentration index value from the facial emotion is in between 50% - 100%, and he/she is also focused.
- Nominally engaged: a student engagement is under this category when the student is focused and the concentration index value from the facial emotion is below 50%.
- Not engaged at all (alert stage): a student engagement is under this category when the student is distracted, i.e., when the output from eye-head movement analysis is 0.

Fig .3 presents a general view of the system working in real time, and it presents the information data to the instructor/teacher. These data, which includes the facial emotions of the learner/student and eye-head movements, can be used to monitor the student/learner in real time while a teacher/instructor/e-learning system is delivering the content.

## 4 Experimental Results and Discussion

The system was tested with 15 students, from different teaching institutions, and ages in the range between 20- 30 years old. Students were requested to see an informative video lecture, which is 1 minute and 56 seconds long, on one general topic on germs and diseases in a normal/traditional class environment. Fig. 3. demonstrates the general view of the system working in real time. The starting frames (0-50) consists of an informative and attractive colorful diagram of a pyramid. In the middle of the video, as illustrated with the question mark ('?') in Fig. 4., the tutor/teacher gives specific instructions about which component will fit in the missing part of the pyramid (the tutor changed this information accordingly).

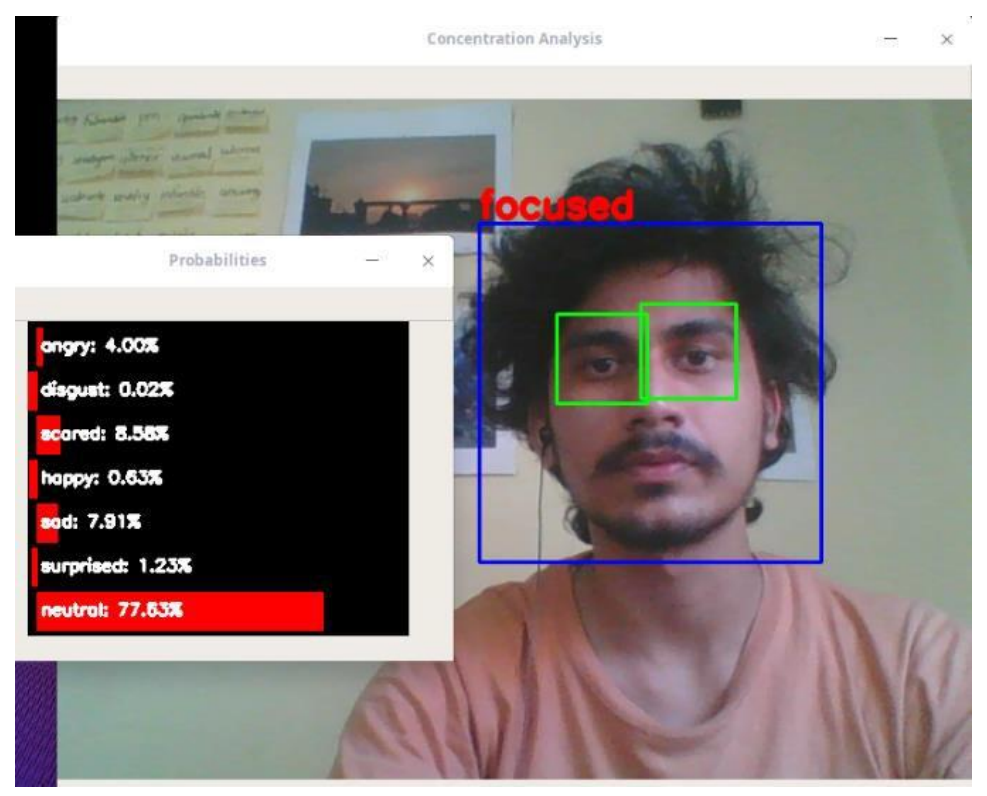

**Fig. 3.** General view of the system working in real time.

The last section of the video consists of almost only verbal information, with this section being shorter than the other sections of video. We have recorded individual videos capturing the facial emotions for each student. At the end of the video presentation, a very simple quiz (3 minutes long) was given to each student. The questions in the quiz have a direct answer in the presented video (i.e., the questions can be easily answered if the student watches the video lecture attentively). Fig. 4. represents a figure which was asked to complete by students. The quiz consisted of the following questions:

i) Which topic did you studied?
   a) Bacteria and Insect
   b) Bacteria Only
   c) Virus Only
   d) Bacteria and Virus

ii) Who is host?
   a) Human
   b) Bacteria
   c) Virus
   d) All

iii) What is the missing part in the Fig. 4?
   a) Host
   b) Virus
   c) Bacteria
   d) None

iv) Bacteria like when the respiratory track is . . .
   a) Cold
   b) Hot
   c) Mild
   d) None

v) What makes the host vulnerable?
  a) When the host has more microbes
  b) When the virus has more microbes
  c) a and b
  d) None

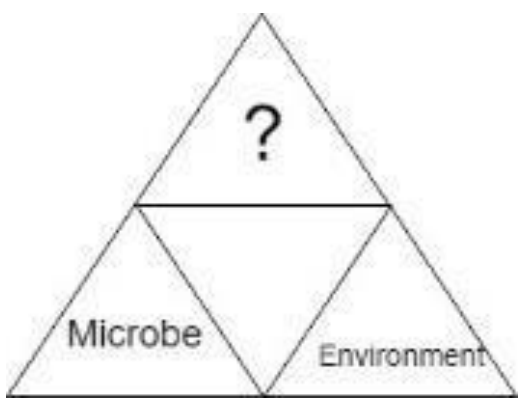

**Fig. 4.** Figure for the quiz.

Table 2 presents the global results of this quiz, along with the partial concentration indexes, for each subject and question. Before we proceed with the analysis of these results, we will explain the procedure used to find the values in this table.

Fig. 5. shows examples of students watching the video, and the corresponding concentration index plots over the time for the full video.

As can be seen from these plots, some students have some intervals with a constant concentration index, while others have fluctuating plots. These fluctuations are due to eye/head movements and negative emotions.

Fig. 6. plots the evolution of the concentration of a student, based solely on the movements of the eyes and head. Since we have used a binary classifier, the concentration index will be either 0 or 100, in percentage. The value 0 signifies that the student is "Distracted", whereas the value 100 signifies that the student is "Focused" (e.g., we can see that, in Fig. 6., a student is "Focused" during the interval between 10th to 70th frames, and from frame 70th to frame 105th the student is "Distracted"). Using algorithmic language, this can be represented as:

if c > 0.5 then
a = 100 (i.e., Focused)
else a = 0 (i.e., Distracted),

where c represents the value given by the Distraction Detector and the concentration in percentage. Note that the value of c is determined as explained in the beginning of subsection III-B above.

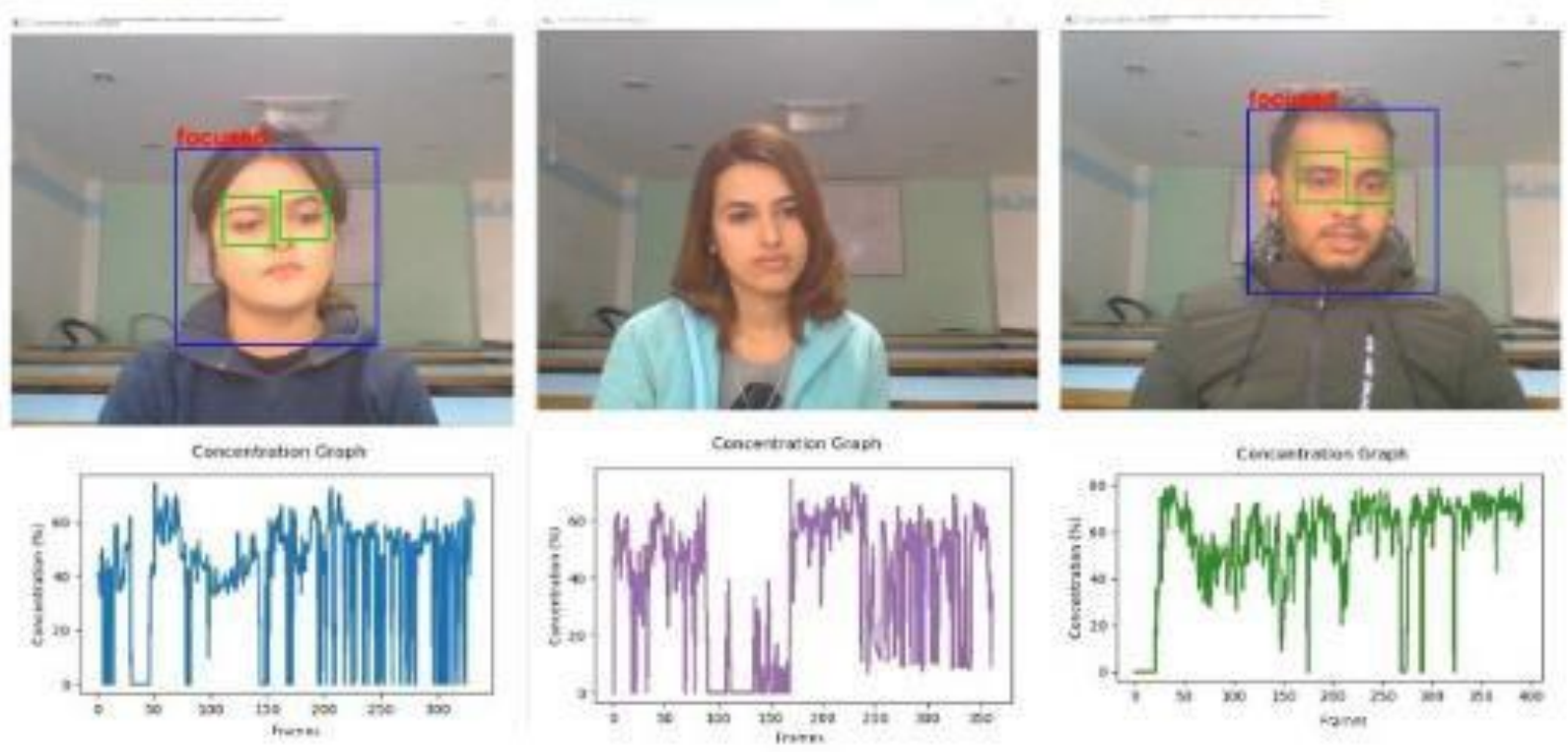

**Fig. 5.** Examples of students watching the video and the corresponding concentration index plots over the complete video.

Fig. 7. plots the evolution of the concentration of a student, based solely on the facial emotions. The concentration index depends upon the seven basic emotions presented above. Each emotion contributes with its own value to the presented concentration index, to extrapolate the concentration value. To determine the concentration in percentage, a, based only on the facial emotion detected, we used the following rules:

if 'neutral' then $CI = (DEP \times 0.9) \times 100$
else if 'happy' then $CI = (DEP \times 0.6) \times 100$
else if 'surprised' then $CI = (DEP \times 0.5) \times 100$
else if 'sad' then $CI = (DEP \times 0.3) \times 100$
else if 'scared' then $CI = (DEP \times 0.3) \times 100$
else if 'angry' then $CI = (DEP \times 0.25) \times 100$
else if 'disgust' then $CI = (DEP \times 0.2) \times 100$
else $CI = 0$ (i.e., Distracted),

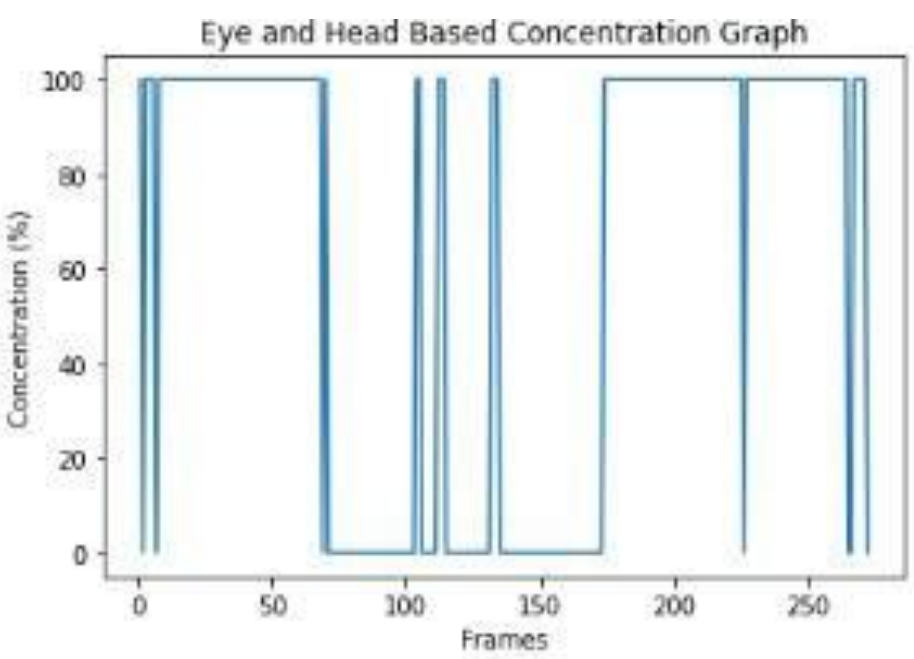

**Fig. 6.** Concentration index based only in the movements of the eyes and head.

The weight values (0.9, 0.6, 0.5, 0.3, 0.3, 0.25, 0.2) have been found experimentally, and presented in Table 1. The DEP score is determined using the machine learning algorithm presented above in section III, then the concentration in percentage, CI, is calculated based on the dominant emotion, explained in subsection III-C. Fig. 7. shows an example where we can be seen that between frames 90 and 120, the student reveals a high concentration level, and between frames 150 and 190 the student shows a low concentration level.

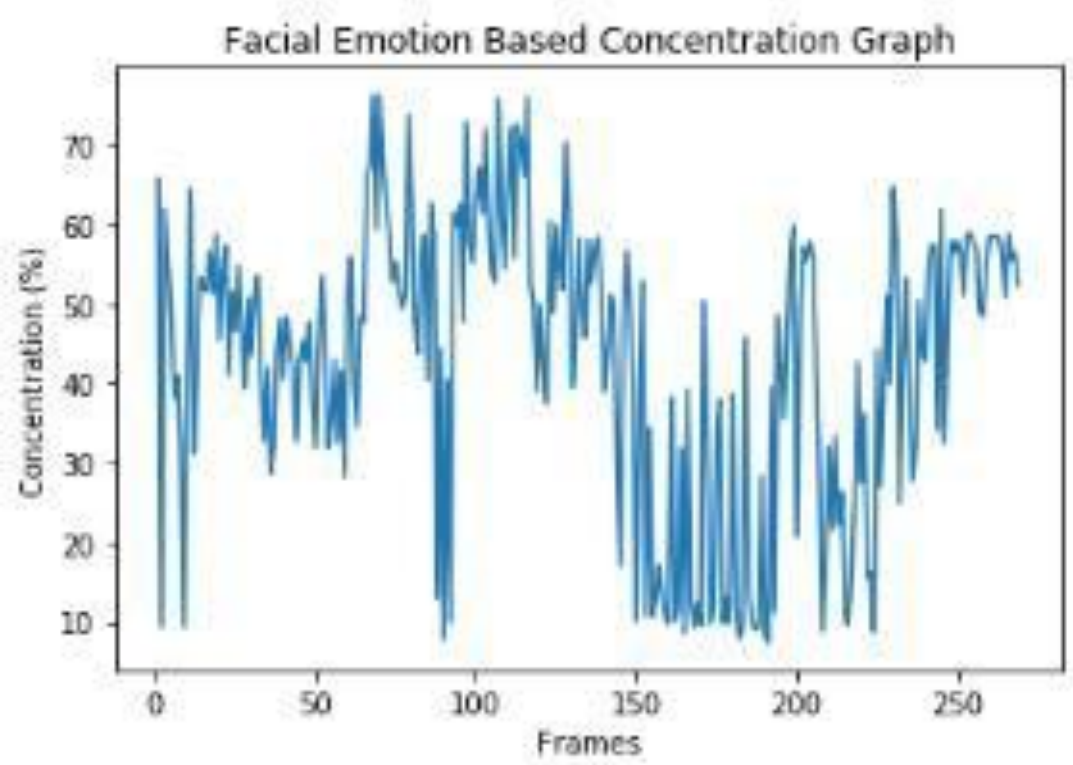

**Fig. 7.** Concentration index based only on the facial emotion's prediction.

Finally, Fig. 8. plots the evolution of the concentration of a student, based on both the facial emotions, and eye and head movements. First, the eye and head movements are analyzed. If this first analysis results in "Focused" state, then facial emotions are predicted to determine the concentration value, using the same rules and weights as presented and explained above. As can be seen in Fig. 8., from frame 12 to frame 47 the student is "Focused", having a concentration index value greater than 45 (in average). Between frames 109 and 140, the student is totally distracted, resulting in a concentration index of 0. Once again, note that if $c < 0.5$, then the student is "Distracted" and an alert can be issued (e.g., the instructor can be notified).

From Table 2, we can see that, out of the 15 students who performed the quiz, all of them have correctly answered question number 3 (which asks the student to find the missing part of the pyramid that was shown in the video) and the resultant mean concentration of the respective question is 60%, which is in "Highly engaged" category. From this result we can conclude that the figure in the e-learning material helps the students to score better, as they have high concentration.

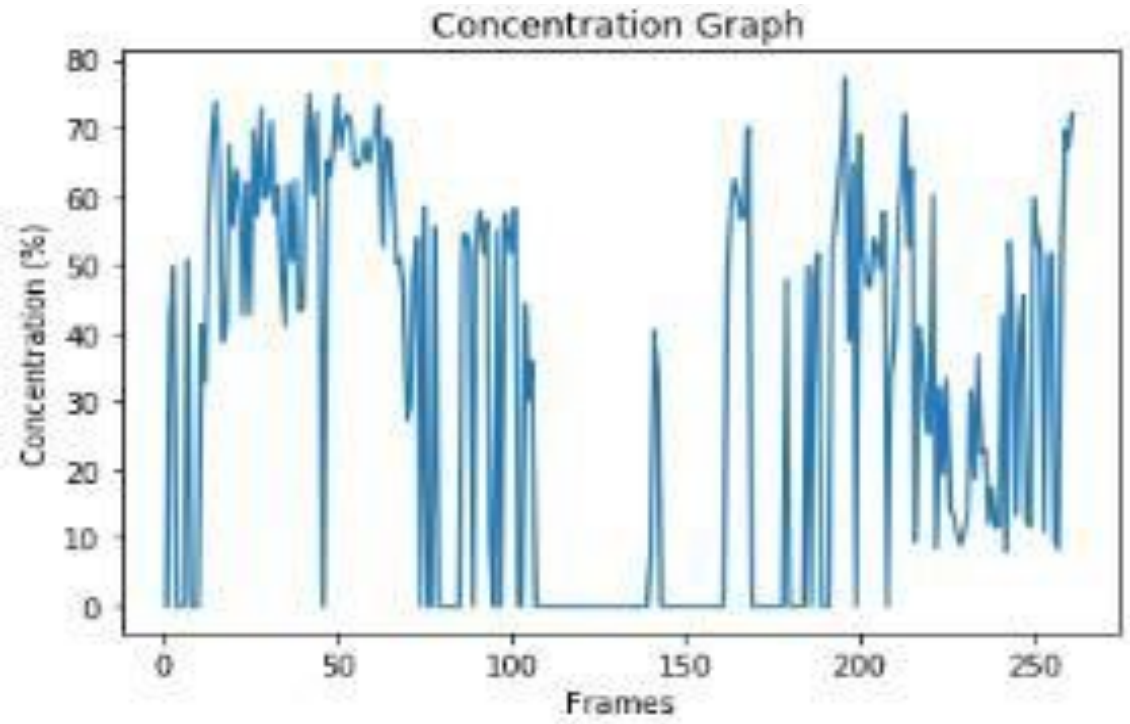

**Fig. 8.** Concentration index including eye, head movements and facial emotions prediction.

**Table 2.** Global Results of the Quiz, along with the partial concentration indexes, for each student and questions

| | Question 1 | | Question 2 | | Question 3 | | Question 4 | | Question 5 | | Total | |
|---|---|---|---|---|---|---|---|---|---|---|---|---|
| Student | Score | CI | Score | CI | Score | CI | Score | CI | Score | CI | Score | CI |
| 1 | 0 | 46.4 | 0 | 34.8 | 1 | 37.6 | 1 | 35.9 | 0 | 28.3 | 2 | 36.6 |
| 2 | 1 | 54.4 | 0 | 56.8 | 1 | 27.1 | 0 | 61.6 | 0 | 56.0 | 2 | 51.2 |
| 3 | 1 | 68.6 | 0 | 64.4 | 1 | 74.1 | 1 | 71.8 | 0 | 68.7 | 3 | 69.5 |
| 4 | 1 | 57.5 | 1 | 46.3 | 1 | 49.7 | 0 | 61.3 | 0 | 49.0 | 3 | 52.8 |
| 5 | 1 | 62.1 | 1 | 60.6 | 1 | 60.3 | 1 | 66.7 | 1 | 60.5 | 5 | 62.0 |
| 6 | 0 | 62.2 | 1 | 52.2 | 1 | 68.1 | 0 | 66.9 | 0 | 54.8 | 2 | 60.8 |
| 7 | 0 | 63.8 | 1 | 71.3 | 1 | 52.9 | 1 | 58.5 | 0 | 56.8 | 3 | 60.7 |
| 8 | 1 | 46.0 | 1 | 1.1 | 1 | 53.9 | 1 | 50.6 | 0 | 27.5 | 4 | 35.8 |
| 9 | 1 | 72.6 | 1 | 71.0 | 1 | 77.1 | 1 | 73.4 | 0 | 61.1 | 4 | 71.0 |
| 10 | 1 | 57.6 | 0 | 47.6 | 1 | 58.0 | 1 | 51.3 | 0 | 47.0 | 3 | 52.3 |
| 11 | 0 | 74.2 | 1 | 69.8 | 1 | 73.2 | 1 | 74.1 | 1 | 72.6 | 4 | 72.8 |
| 12 | 0 | 71.1 | 1 | 68.1 | 1 | 68.9 | 1 | 65.5 | 1 | 74.3 | 4 | 69.6 |
| 13 | 1 | 64.3 | 1 | 49.2 | 1 | 62.5 | 0 | 63.5 | 0 | 63.7 | 3 | 60.6 |
| 14 | 1 | 68,8 | 1 | 70.1 | 1 | 66.5 | 1 | 69.1 | 1 | 65.1 | 5 | 67.9 |
| 15 | 1 | 67.6 | 1 | 61.8 | 1 | 72.4 | 0 | 61.3 | 1 | 63.1 | 4 | 65.2 |
| Mean | 0.7 | 62.5 | 0.7 | 55.0 | 1 | 60.2 | 0.7 | 62.1 | 0.3 | 56.6 | 3.4 | 59,3 |
| Mode | 1 | --- | 1 | --- | 1 | --- | 1 | --- | 0 | --- | 3 | --- |
| StD | 0.5 | 8.4 | 0.4 | 17.9 | 0 | 13.7 | 0.5 | 9.7 | 0.5 | 13.5 | 1.0 | 11.2 |

Levels during that learning period. Additionally, we can see that 92% of the students who scored 3 and above are also in the "Highly engaged" category (between 50-100%), which signifies that the score is correlated with their respective concentration value. Another finding reveals that the majority of students were not able to answer question number 5 (which is in the last section of video lecture). This might have happened due to the stress they feel with lots of new information, and the relatively short period of time they had have to absorb all the contents, and at the end they get saturated.

While the majority of the results are aligned with the concentration index values determined by the system presented here, this is not the case for the scored values for students 2, 6 and 8.

A detailed analysis of the entire videos of these students revealed that our system was unable to deal with the problem of "face occlusion". In our case, face occlusion is a state in which the face of the student is partially covered by his/her hand/s or any other object (such as glasses). As mentioned by [25], one of the major face occlusion is "hand to face gesture in which face is covered by hand". In the particular case of face recognition systems, this problem results in a lowering of the performance rate of these systems [26]. Fig. 9. shows examples of student 2 and 6 partially covering their faces with their hands. As can be seen from the images, the students are focused because their eye-head movement is positive, i.e., they are effectively watching the content in the screen. However, the system is unable to extract their facial emotions completely, which results in a low concentration index. As a consequence, the system regards them as students with low concentration level, even if they are watching the content with high attention. In some particular cases, although the system gives a relative high value for the concentration, contrasting with the low marks they achieved in the quiz, these results, as explained by the students, could be because of the stress of their approaching board exams. Concerning the results of student 8, our system was unable to correctly find the concentration index due to the fact that this student wore glasses, and the concentration values given by the system were fluctuating. It is our aim to correct this effect in the future versions of the system presented here.

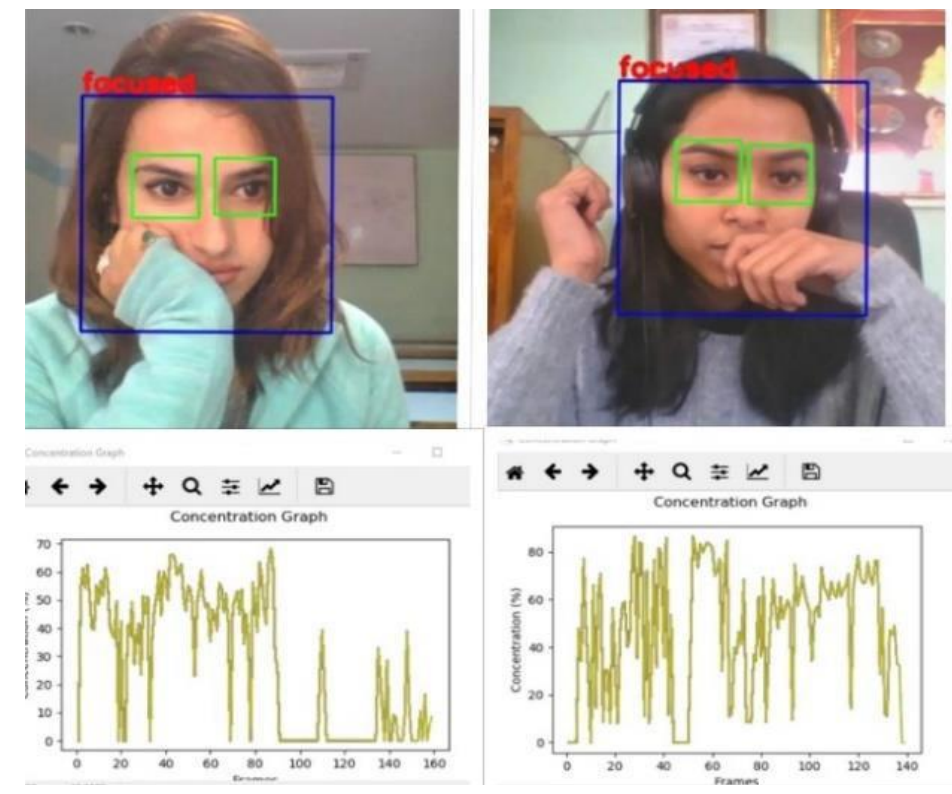

**Fig. 9.** Facial recognition error due to hand gesture.

## 5 Conclusion and Future Work

With the increase of distance learning, and e-learning environments in particular, having a system capable of determining students' engagement is of primordial importance and one of the biggest challenges both for teachers, researchers and policy makers.

Here, we presented a new approach of a system to detect the engagement level of the students. The system uses only the information provided by the built-in web-camera present in typical laptop computer. Our system uses the images grabbed by the camera to extract information about the movements of the eyes and head, and combines this information with the facial emotions, also retrieved from these images, to produce a concentration index. The presented system produces three classes of engagement: "very engaged", "nominally engaged" and "not engaged at all". The system proposed here was designed to work in real time.

By including the facial emotions information reflected by students about the learning topic, which includes the seven typical emotions, a teacher/instructor/learning management system will have live feedback, hence helping the system/teacher/instructor to automatically adapt the learning contents to the needs of the students. This will definitively contribute to dynamically enrich the learning environment and, hence, improve the performance of the students.

We have tested our system with fifteen students in a typical e-learning scenario, and the results show that the system correctly identifies each period of time where students were "very engaged", "nominally engaged" and "not engaged at all". Additionally, the results also show that the students with best scores also have higher resultant concentration index. In the future, we want to merge the information currently provided by our system with the information retrieved with the help of other sensors, such as heart rate, EEG signals, and oxygen level, among other. We are also working to change to 3D facial expressions detection, because they better facilitate an examination of the fine structural changes inherent to spontaneous expressions. However, this will pose additional difficulties and the need to use other capturing cameras, besides the laptop typical built- in web-camera